\documentclass{style/naturep}

\usepackage{amssymb}
\usepackage{amsmath}
\usepackage{graphicx}
\usepackage{algorithmicx}
\usepackage{algorithm}
\usepackage{adjustbox}
\usepackage{bm}
\usepackage{tabularx}
\usepackage{booktabs}
\usepackage{multirow}
\usepackage{makecell}
\usepackage{pifont}
\usepackage[table,x11names]{xcolor}
\usepackage{xurl}
\usepackage{enumitem}
\setlist[itemize]{noitemsep, nolistsep}
\usepackage[noend]{algpseudocode}
\usepackage{bbm}
\usepackage{lineno}
\usepackage[margin=0.6in]{geometry}
\usepackage{ragged2e}
\usepackage{setspace}
\usepackage{longtable}
\usepackage{hyperref}
\usepackage{anyfontsize}
\usepackage{float}
\usepackage{placeins}
\usepackage{caption}
\usepackage{subcaption}
\usepackage{enumitem, kantlipsum}
\usepackage{xspace}
\usepackage{colortbl}
\usepackage{afterpage}

\definecolor{maroon}{cmyk}{0,0.87,0.68,0.32}
\definecolor{gray}{rgb}{0.3,0.3,0.3}
\definecolor{LightGray}{gray}{0.9}

\newcommand{\ours}{\textsc{STORM}\xspace}
\newcolumntype{Y}{>{\centering\arraybackslash}X}

\newcommand\Heading[1]{
  \noindent\textbf{\Large{#1}}
}

\newcommand\heading[1]{
  \noindent\textbf{\large{#1}}
}

\title{\begin{flushleft}{\begin{spacing}{1}
   A Multimodal Foundation Model of Spatial Transcriptomics and Histology for Biological Discovery and Clinical Prediction
\end{spacing}}\end{flushleft}}

\makeatletter
\let\saved@includegraphics\includegraphics
\AtBeginDocument{\let\includegraphics\saved@includegraphics}

\makeatother

\begin{document}

\maketitle
\vspace{-20mm}
\begin{spacing}{1.4}
\noindent
Jinxi Xiang$^{1\boldsymbol{\ddag}}$,
Siyu Hou$^{2\boldsymbol{\ddag}}$,
Yuchen Li$^{1\boldsymbol{\ddag}}$,
Ryan Quinton$^{3}$,
Xiaoming Zhang$^{4}$,
Feyisope Eweje$^{1,7}$,
Xiangde Luo$^{1}$,
Yijiang Chen$^{1}$,
Zhe Li$^{1}$,
Colin Bergstrom$^{3}$,
Ted Kim$^{1}$,
Sierra Willens$^{3}$,
Francesca Maria Olguin$^{3}$,
Matthew Abikenari$^{5}$,
Andrew Heider$^{1}$,
Sanjeeth Rajaram$^{3}$,
Joel Neal$^{3}$,
Maximilian Diehn$^{1}$,
Xiang Zhou$^{2*}$,
Ruijiang Li$^{1,6*}$

\end{spacing}

\vspace{-7mm}
\begin{spacing}{1.4}
\begin{affiliations}
\item Department of Radiation Oncology, Stanford University School of Medicine, Stanford, CA, USA
\item Department of Statistics and Data Science, Yale University,  New Haven, CT, USA
\item Department of Medicine (Oncology), Stanford University School of Medicine, Stanford, CA, USA
\item  Department of Pathology, Stanford University School of Medicine, Stanford, CA, USA
\item Department of Neurosurgery, Stanford University School of Medicine, Stanford, CA, USA
\item Stanford Institute for Human-Centered Artificial Intelligence, Stanford, CA, USA
\item Perelman School of Medicine at the University of Pennsylvania
, Philadelphia, PA, USA
 \\ $\boldsymbol{\ddag}$ Equal contribution
 \\\textbf{*} Correspondence to: Ruijiang Li (rli2@stanford.edu), Xiang Zhou (xiang.zhou.xz735@yale.edu)
\end{affiliations}
\end{spacing}

\noindent This manuscript is a work in progress; further updates and revisions will be posted as they become available.

\noindent Project Page: \url{https://storm-web-demo.vercel.app/}

\vspace{10mm}

\clearpage

\Heading{Abstract}
\begin{spacing}{1.38}
\noindent
Spatial transcriptomics (ST) enables gene expression mapping within anatomical context but remains costly and low-throughput. Hematoxylin and eosin (H\&E) staining offers rich morphology yet lacks molecular resolution. We present \textbf{\ours} (\textbf{S}patial \textbf{T}ranscriptomics and hist\textbf{O}logy \textbf{R}epresentation \textbf{M}odel), a foundation model trained on 1.2 million spatially resolved transcriptomic profiles with matched histology across 18 organs. Using a hierarchical architecture integrating morphological features, gene expression, and spatial context, STORM bridges imaging and omics through robust molecular--morphological representations. STORM enhances spatial domain discovery, producing biologically coherent tissue maps, and outperforms existing methods in predicting spatial gene expression from H\&E images across 11 tumor types. The model is platform-agnostic, performing consistently across Visium, Xenium, Visium HD, and CosMx. Applied to 23 independent cohorts comprising 7,245 patients, STORM significantly improves immunotherapy response prediction and prognostication over established biomarkers, providing a scalable framework for spatially informed discovery and clinical precision medicine.
\end{spacing}

\vspace{1em}

\begin{spacing}{1.35}
\clearpage
\Heading{Introduction}

\noindent
Spatial transcriptomics (ST) is a powerful technology that enables high-resolution, spatially resolved gene expression profiling within the native tissue context\cite{rao2021exploring, moses2022museum, bressan2023dawn}. ST has transformed our understanding of human biology in health and disease by revealing novel insights about disease mechanisms and therapeutic targets\cite{ravi2022spatially, chen2024spatial, harnik2024spatial, sun2024spatial, yayon2024spatial, jia2025spatial, pei2025spatial}. Despite its vast potential, however, broad applications of ST have been hampered by a number of technical challenges such as high cost, complexity, and low throughput, restricting its use to small-scale, research settings\cite{walker2024insights, jain2024spatial, gong2024spatial}. These constraints severely hinder its translation and deployment in clinical settings.

\noindent
Histology with hematoxylin and eosin (H\&E) staining, on the other hand, is a widely used tool in clinical diagnosis. It contains rich morphological information that is reflective of the underlying molecular programs. Recent studies have demonstrated that histological patterns can be computationally decoded using deep learning to infer molecular traits, including spatially resolved gene expression\cite{he2020integrating,zeng2022spatial,xie2023spatially, fu2025spatial}. However, these studies are limited by their narrow scope with a focus on specific indications and the use of small datasets. In addition, the models require task-specific training that necessarily limits their generalizability across broad applications.

\noindent
Here, we present a new multimodal foundation model for integrating spatial transcriptomics and histology (\ours). The model is pretrained on 1.2 million spots with spatially resolved transcriptomic profiles and matched {high-resolution} histology images across 18 tissue types. Different from previous approaches focused on learning on an individual-spot basis, we design a novel hierarchical model architecture that not only learns spot-level features but also incorporates spatial context information from its neighborhood. This ``spatially aware" design is crucial for modeling the spatial organization of tissue architectures.

\noindent
Through comprehensive evaluation, we show that \ours improves spatial domain discovery given multi-modal ST and H\&E data, generating coherent and biologically meaningful maps of tissue compartments. In addition, \ours demonstrates substantially superior accuracy over previous methods for H\&E-based spatial gene expression prediction across eleven tumor types.
The framework is platform-agnostic, providing robust performance across both spot-level (Visium) and single-cell resolution technologies (Xenium, Visium HD, and CosMx).
Further, we apply \ours to predict clinical outcomes across {23} independent real-world cohorts with a total of {7,245} patients.

\clearpage
\Heading{Study Overview}

\noindent
The development of \ours addresses the need for a unified framework capable of capturing complex molecular–morphological relationships across diverse human tissues. We trained \ours on a large-scale, diverse spatial ST-H\&E dataset containing \textbf{1.2 million spatially resolved spots} with matched high-resolution images from 632 tissue sections across 18 human organs. The scale and diversity of this dataset are critical for capturing representative patterns that are generalizable across heterogeneous tissue types.

\noindent
To effectively integrate these multimodal signals, we designed a spatially aware, hierarchical model architecture. The model first extracts local morphological and transcriptomic representations using dedicated spot encoders. These features are then integrated by a spatial encoder that incorporates context information from neighboring spots and their spatial locations. To optimize this architecture, we implemented a self-supervised, multimodal masked pretraining strategy that reconstruct missing biological information by randomly masking tissue spots. This objective forces \ours to learn molecular–morphological representations in spatial domain. In this way, \ours develops a robust, holistic understanding of spatial biology.

\noindent
We conducted comprehensive evaluations and benchmarking of \ours\ in diverse biological and translational applications. First, for spatial domain discovery, \ours\ outperforms both classical methods and recent foundation models\cite{chen2025visual, wang2025scgpt, dong2022deciphering, long2023spatially, hu2021spagcn}, producing more coherent, biologically meaningful tissue compartment maps.
Second, in ST prediction from H\&E images, \ours\ substantially surpasses H\&E–ST models\cite{xie2023spatially, min2024multimodal, hoang2024deep, yang2024spatial} and pathology models \cite{xu2024whole, chen2024towards, vorontsov2024foundation} across {eleven} tumor types, opening the door for scalable, routine H\&E-based virtual ST analysis without specialized assays.
Finally, we assessed the translational value of \ours for predicting clinical outcomes from H\&E across {23} real-world cohorts, including {pancancer prognosis prediction}, and immunotherapy response prediction in melanoma, lung, gastroesophageal, kidney, bladder, and endometrial cancers. We show that by integrating H\&E and virtual ST inferred by \ours, the multimodal approach consistently outperforms either modality alone and competing methods, demonstrating significant translational value for precision oncology.

\begin{figure*}[!h]
\centering
\includegraphics[width=0.97\linewidth]{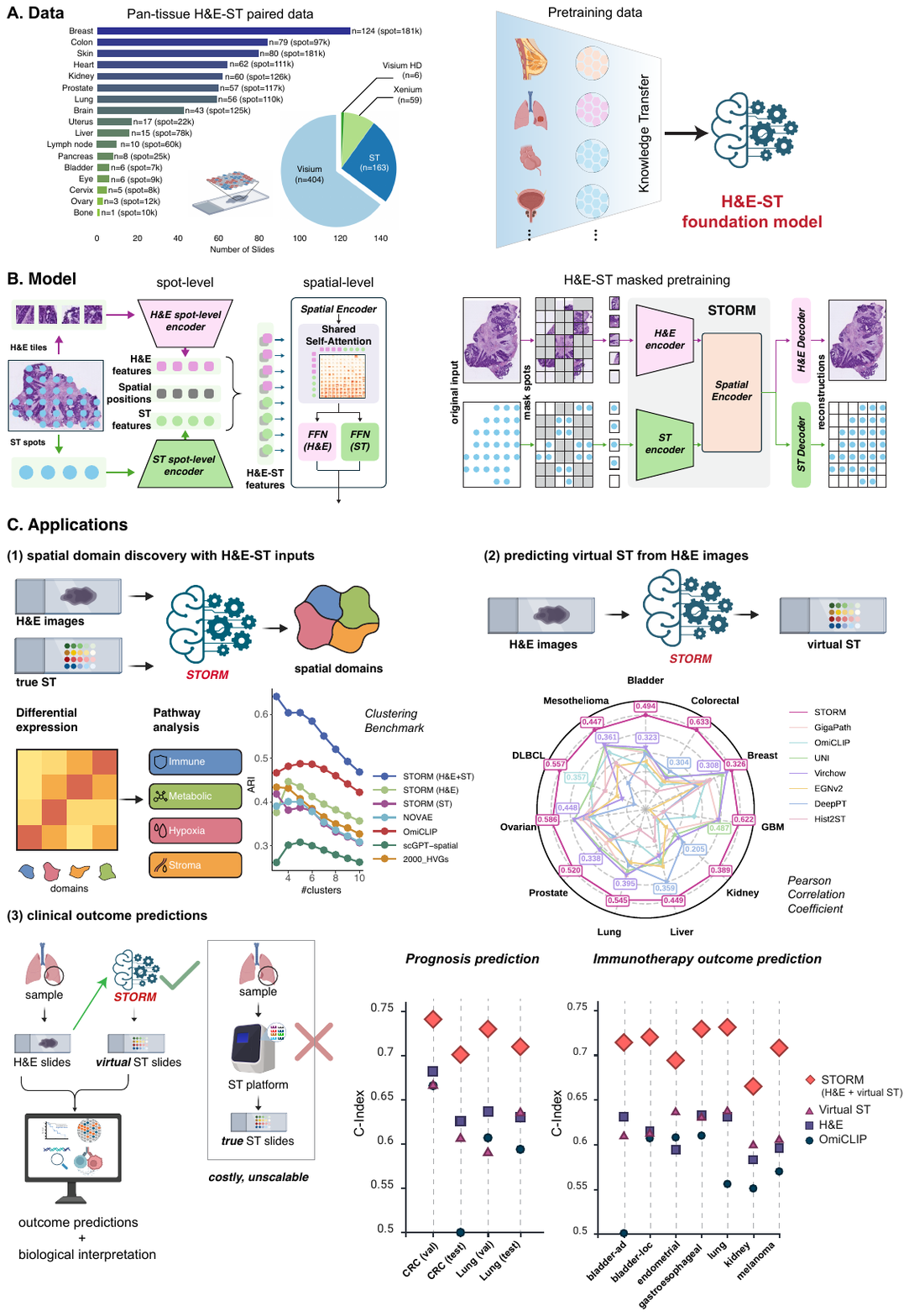}
\end{figure*}
\clearpage 
\begin{figure*}
\caption{\textbf{Study overview of the proposed ST-H\&E foundation model, \ours.}  
\textbf{(A)} We trained \ours with a large-scale human pan-tissue dataset comprising 632 paired ST–H\&E slides across four major spatial transcriptomics platforms and 18 organs, totaling $\sim$1.2 million {high-resolution} tissue spots. 
\textbf{(B)} Model architecture and pretraining strategy. \ours\ employs a hierarchical design: first, spot-level foundation models are used to process each spot independently; Second, the image and ST features for each spot, together with the spatial coordinates, are then integrated by a spatial encoder which not only fuses H\&E–ST information but also preserves \textit{spatial context}. \ours is fully compatible with the heterogeneous ST platform, accommodating variations in spatial resolution and gene panels.  
\textbf{(C)} After pretraining, \ours\ can be applied to various tasks with minimal or no further fine-tuning. First, on comprehensive benchmarks for spatial domain discovery with H\&E–ST inputs, \ours\ substantially outperforms OmiCLIP\cite{chen2025visual} and other ST foundation models. Second, for virtual ST prediction from H\&E images across {11} organ types, \ours\ surpasses other H\&E–ST models  and H\&E-only foundation models. Finally, in large-scale real-world patient cohorts, we demonstrate the translational potential of \ours\ for clinical outcome prediction by integrating H\&E and virtual ST, thereby eliminating the need for costly ST. \ours\ markedly improves predictive accuracy across {23} independent cohorts with {7,245}, including {pan-cancer prognosis prediction}, and immunotherapy response prediction in melanoma, bladder (advanced and localized), endometrial, gastroesophageal, lung, and kidney cancers. 
}
\label{fig:fig1-Overview}
\end{figure*}

\Heading{Results}

\heading{\ours improves multimodal spatial domain discovery}

\noindent
To evaluate the multi‐modal capability of \ours\ in spatial domain discovery, we utilized a Visium ST dataset\cite{dawo_2025_14620362} with kidney cancer (KC) and lung cancer (LC) samples, each paired with high‐resolution H\&E‐stained histopathology images and pathologist annotations (Fig. \ref{fig:fig2-spatial_domain}A). Spatial domain detection was performed with multimodal H\&E-ST embeddings from \ours, which is then benchmarked against multiple baselines, including using the 2,000 highly variable genes (HVGs), ST foundation models (NOVAE\cite{blampey2024novae}, scGPT spatial\cite{wang2025scgpt}), and a H\&E-ST foundation model, OmiCLIP\cite{chen2025visual}.

\noindent
In Fig. \ref{fig:fig2-spatial_domain}B, \ours\ (H\&E + ST) achieved clear separation of key domains, including tumor (TUM), normal (NOR), tertiary lymphoid structures (TLS), and  immune infiltration (INFL), in the UMAP embedding. Even with ST or H\&E input alone, \ours\ is able to yield distinct and biologically coherent clusters. By contrast, the conventional 2,000 HVGs baseline failed to resolve these microenvironments, resulting in substantial domain overlap.
Across eight samples (three KC, five LC), \ours\ achieved the highest concordance with ground truth annotations: in KC, average mean NMI(Normalized Mutual Information)/ARI(Adjusted Rand Index) = 0.53/0.40 versus OmiCLIP's 0.41/0.32 (other methods $<0.30/<0.20$); in LC, NMI/ARI = 0.62/0.60 versus OmiCLIP's 0.58/0.49 (Figs. \ref{fig:fig2-spatial_domain}D,E). Also, \ours\ yielded superior spatial homogeneity (HOM = 0.67 in KC, 0.68 in LC) and lowest abnormal spot percentages (PAS = 0.04 and 0.02, respectively). When adjusting the number of clusters ($K=3\text{--}10$), \ours's embeddings maintained markedly higher stability than other approaches (Fig. \ref{fig:fig2-spatial_domain}F), validating the robustness of its multimodal representations. Similar results were observed on eight breast cancer samples, where \ours consistently ranked first across all metrics.

\noindent
{We further compared \ours\ with established spatial domain detection methods, including STAGATE\cite{dong2022deciphering}, GraphST\cite{long2023spatially}, and SpaGCN\cite{hu2021spagcn}, as well as a baseline of spatially variable genes (SVGs). Across all evaluated datasets, \ours\ consistently achieved higher agreement with pathologist annotations, as measured by NMI and ARI, than these spatial domain--specific approaches and the SVG-based baseline. These performance differences reflect fundamental methodological distinctions. Conventional spatial domain detection methods primarily rely on gene expression profiles combined with spatial smoothing or graph-based regularization, which can enforce local spatial continuity and, in some cases, yield slightly improved spatial coherence metrics (e.g., lower PAS) than other foundation models. However, such unimodal designs limit their capacity to capture complex tissue architecture and morphological context. In contrast, \ours\ integrates histopathology images with spatial transcriptomics in a unified multimodal embedding, enabling richer representation of spatial, morphological, and molecular heterogeneity.
}

\noindent
We illustrate the biological insights gained through \ours for understanding spatial organization and heterogeneity of the tumor microenvironment. Specifically, in KC2, the ground-truth annotations by pathologists only provide a coarse-grained ``tumor" label. However, \ours delineated three spatially distinct subdomains (Tumor Domains 1--3; Fig. \ref{fig:fig2-spatial_domain}G) within the histologically annotated ``tumor" region, with Domains 2 and 3 forming a contiguous cluster. To investigate the driving factors behind this result, we performed differential expression analysis in each domain against its adjacent regions, identifying a set of genes significantly up-regulated in Domain 3 (Fig. \ref{fig:fig2-spatial_domain}H). Among the top genes were FABP7 and RARRES2(Chemerin), which promote fatty acid metabolic reprogramming facilitating lipid uptake and storage to support tumor growth and survival in ccRCC\cite{zhou2015overexpression, koundouros2020reprogramming, tan2023fatty}; IGFBP3, a key modulator of IGF-I--dependent proliferation and survival\cite{cheung2004roles}, and coagulation factors FGB/FGG together with complement C3, implicating activation of the tumor coagulome and complement cascade in microenvironmental remodeling and metastatic facilitation\cite{palumbo2000fibrinogen, ajona2019complement, galmiche2022coagulome, zhang2019role}. Spatial feature plots (Fig. \ref{fig:fig2-spatial_domain}I) confirm the region-specific expression of these markers in Domain 3. Altogether, these markers reveal a tumour subregion characterized by heightened metabolic flux, chemotactic signalling, inflammatory activation, and stress adaptation, hallmarks of an aggressive phenotype.

\noindent
Gene set enrichment against the MSigDB hallmark collection\cite{liberzon2015molecular} further revealed coordinated activation of multiple pathways in Domain 3, including oxidative phosphorylation, glycolysis, MYC targets, and mTORC1 signalling---indicative of metabolic plasticity---alongside hypoxia and p53 pathways, reflecting adaptation to microenvironmental stress; EMT and adipogenesis programs, driving matrix remodeling and lipid storage; and TNF-$\alpha$/NF-$\kappa$B and IL-2/STAT5 signalling, representing inflammatory and immune-regulatory circuits (Fig. \ref{fig:fig2-spatial_domain}J). Together, these data portray Domain 3 as a central hub integrating diverse metabolic routes, stress-resilience mechanisms, invasive programs, and inflammatory signaling to fuel proliferation and metastatic potential. To further validate the results and explore their clinical implications, we performed independent survival prediction on the external TCGA-KIRC ($n=534$) project using the top 10\% of highly expressed differential genes in Domain 3 (approximately 40 genes; Fig. \ref{fig:fig2-spatial_domain}K). In the TCGA-KIRC cohort, patients with high expression of these markers exhibited significantly shorter progression-free intervals, confirming poorer prognosis and substantiating our functional inference for Domain 3.

\noindent
We further elucidate the immunological heterogeneity among kidney cancer samples by using KC3, which features extensive immune infiltrate regions and tertiary lymphoid structures (TLS). We reclustered the \ours embeddings into ten distinct domains (Fig. \ref{fig:fig2-spatial_domain}L), which broadly recapitulated manual annotations while providing finer subdivisions. For example, at a four-cluster resolution (Fig. \ref{fig:fig2-spatial_domain}C), TLS and INFL regions coalesced into a single group; while at ten domains, TLS and INFL regions were separated, and the INFL region split into two subtypes. As demonstrated in Fig. \ref{fig:fig2-spatial_domain}, domain INFL2 was characterized by a high abundance of plasma cells, whereas domain INFL1 exhibited elevated levels of T follicular helper (Tfh) cells, resting NK cells, and activated dendritic cells. The immune microenvironment in domain INFL2 suggests a plasma cell--driven humoral response, potentially reflecting a mature anti-tumor immune phase. These tumor-reactive plasma cells may contribute to local IgG production, antigen presentation, and Fc-mediated T cell recruitment. In contrast, domain INFL1 displays a coordinated innate--adaptive activation signature, with prominent Tfh cells and activated dendritic cells---key components of TLS and efficient antigen presentation. Although NK cells remain in a resting state, they still can modulate myeloid function and initiate IFN-$\gamma$--dependent immune cascades. The marginal enrichment of M0 macrophages, which represent a non-polarized state, may reflect a transitional microenvironment poised to polarize toward either pro-inflammatory (M1) or immunosuppressive (M2) phenotypes---both relevant in ccRCC\cite{chevrier2017immune}.

\begin{figure*}[!h]
\centering
\includegraphics[width=0.9\linewidth]{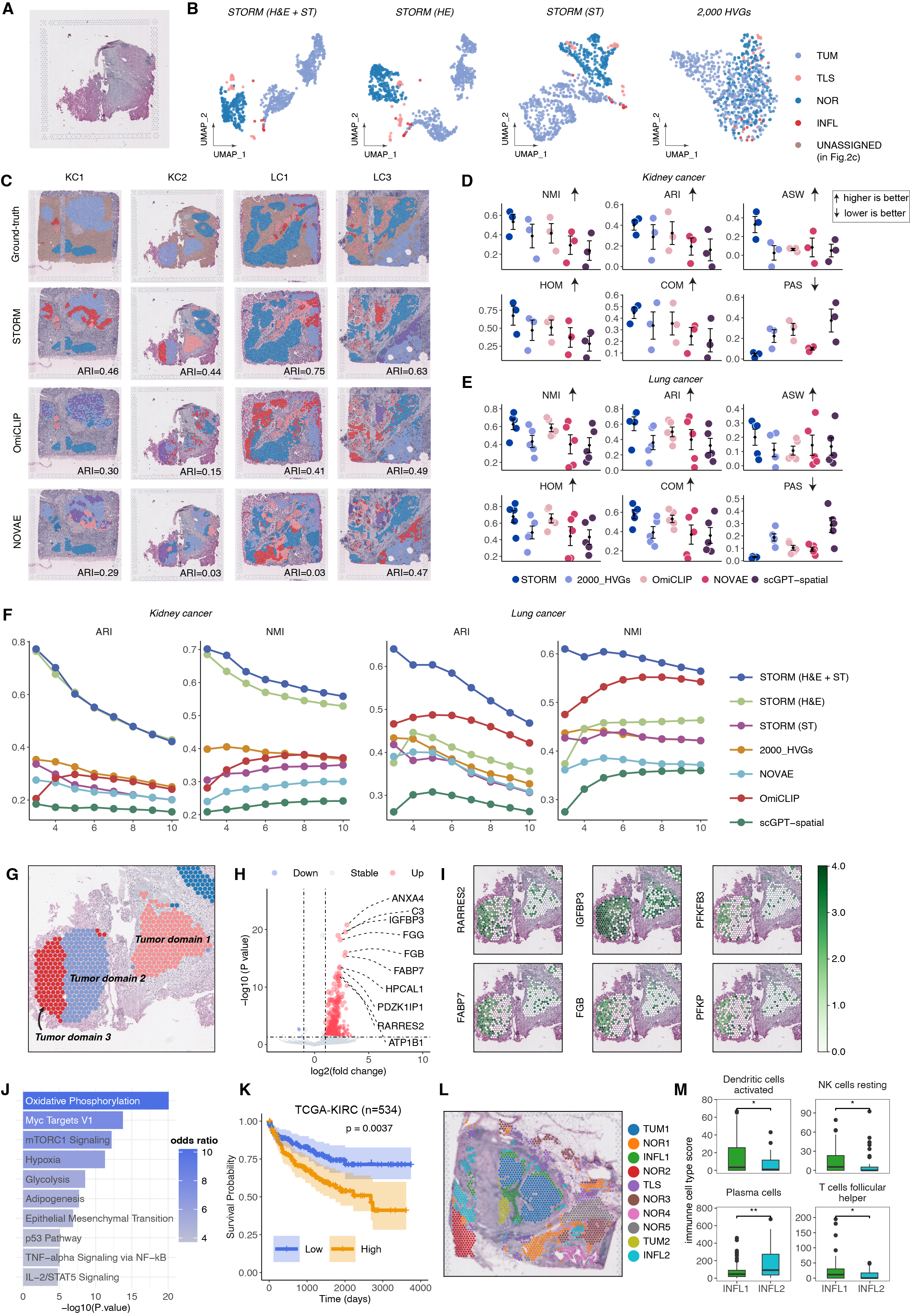}
\end{figure*}
\clearpage
\begin{figure*}[!h]
\caption{
\textbf{Multimodal H\&E-ST spatial domain discovery with \ours.}
\textbf{(A)} Evaluation setup using a Visium ST dataset\cite{dawo_2025_14620362} with kidney cancer (KC) and lung cancer (LC) samples, each paired with H\&E histology and pathologist annotations.
\textbf{(B-C)} Spatial domain detection using multimodal H\&E--ST embeddings of \ours outperforms the classical HVGs+PCA pipeline and foundation models (e.g., OmiCLIP), clearly separating tumor (TUM), normal (NOR), tertiary lymphoid structures (TLS), and infiltrate (INFL) regions.
\textbf{(D--F)} Across eight independent samples, \ours achieves the highest concordance with ground truth (NMI/ARI) and superior spatial homogeneity, with robust performance across clustering resolutions.
\textbf{(G--K)} In KC2, \ours identifies three tumor subdomains, with Domain 3 enriched for metabolic, inflammatory, and invasive programs (e.g., FABP7, RARRES2, IGFBP3, FGB/FGG, C3) and hallmark pathways linked to aggressive phenotypes, validated in TCGA‐KIRC as predicting poorer prognosis.
\textbf{(L--M)} In KC3, fine-resolution clustering resolves TLS and infiltrates into distinct subtypes: INFL2, enriched in plasma cells and humoral immunity, and INFL1, enriched in T follicular helper cells, NK cells, and dendritic cells, indicating distinct immune activation states and interaction patterns.
}
\label{fig:fig2-spatial_domain}
\end{figure*}

\heading{\ours predicts spatial transcriptomics from H\&E images}

\noindent
We evaluated \ours\ for predicting spatial transcriptomics directly from H\&E images. This task requires reconstructing complex gene expression programs that reflect diverse tissue architecture contexts.
We benchmarked \ours\ against {13} state-of-the-art methods from three categories:
(1) \textbf{task-specific H\&E-to-ST models} (BLEEP\cite{xie2023spatially}, mclSTExp\cite{min2024multimodal}, {DeepPT\cite{hoang2024deep}, EGNv1\cite{yang2023exemplar}, EGNv2\cite{yang2024spatial}, Hist2ST\cite{zeng2022spatial}, ST-Net\cite{he2020integrating}, HisToGene\cite{Pang2021LeveragingII}, TCGN\cite{xiao2024transformer})};
(2) \textbf{pathology image foundation models} (GigaPath\cite{xu2024whole}, UNI\cite{chen2024towards}, Virchow\cite{vorontsov2024foundation}); and
(3) the \textbf{H\&E-ST foundation model} OmiCLIP\cite{chen2025visual}.
Model performance was quantified using the Pearson correlation coefficient (PCC) at the gene level between the predicted and measured gene expression under 500 Highly Variable Genes (HVGs).
For this task, we used 8 datasets consisting of 142 tissue sections with a total of 85,236 ST spots and matched H\&E images across eight tumor types. We randomly split the slides with a 4:1 ratio for fine-tuning and validation, and importantly, none of the validation sets were used for pretraining of \ours.

{
\noindent
\ours\ consistently and substantially outperformed all existing baselines in reconstructing spatial gene expression (Fig.~\ref{fig:fig3-he2st_main}A). Across the eight tumor types, \ours\ achieved the highest median PCC in every dataset, demonstrating superior generalization capabilities.
For example, in colorectal cancer, \ours\ achieved a median PCC of $0.633$, representing a $>100\%$ relative improvement over the second-best method, DeepPT ($0.304$), and outperforming the leading pathology foundation model Virchow ($0.266$). Similarly, in GBM, \ours\ ($0.622$) surpassed the strong baseline Virchow ($0.487$) by nearly 28\%.
Notable performance gains were also observed in bladder ($0.494$ vs. UNI $0.323$; $+53\%$), lung ($0.545$ vs. UNI $0.395$; $+38\%$), and prostate ($0.520$ vs. UNI $0.338$; $+54\%$).
Even in challenging tissue contexts such as kidney and breast, where baseline performance was markedly lower (e.g., kidney baselines $<0.21$), \ours\ maintained robust predictive power ($0.389$ in kidney, $+90\%$ vs. DeepPT).

\noindent
Comparative analysis of the baseline categories reveals distinct performance stratifications. Task-specific models generally exhibited high variance and struggled to generalize across the pan-cancer cohort. Within this category, DeepPT and EGNv2 emerged as the top performers, consistent with recent benchmarks\cite{wang2025benchmarking}, yet their success was often restricted to isolated organs (e.g., TCGN in liver, $0.375$; DeepPT in colorectal, $0.304$). This limitation is likely attributed to the lack of large-scale pretraining leveraged by \ours, GigaPath, UNI, and Virchow. Conversely, pathology foundation models (Virchow, UNI) consistently ranked as the second-best performers, validating the utility of large-scale morphological pretraining. However, \ours\ significantly surpassed these pathology models, demonstrating that H\&E-ST multimodal pretraining is essential to accurately reconstruct the fine-grained mapping between histology and transcriptomics.}

\noindent
{
Next, we evaluated the accuracy of whole-transcriptome predictions, focusing on the top genes ranked by PCC (Fig.~\ref{fig:fig3-he2st_main}B, C). We benchmarked performance against DeepPT (the best task-specific model), OmiCLIP (the existing H\&E-ST foundation model), and Virchow (the best baseline of all). For the top 500 genes, \ours\ demonstrated superior predictive fidelity, surpassing all baseline methods across all eight organ types.
The performance gains were particularly decisive in several lineages, with relative improvements over Virchow of $+83\%$ in colorectal ($0.623$ vs. $0.340$), $+47\%$ in prostate ($0.573$ vs. $0.390$), and $+30\%$ in bladder ($0.643$ vs. $0.493$). In the kidney cohort, \ours\ outperformed Virchow by $+54\%$ ($0.426$ vs. $0.277$) and significantly exceeded both DeepPT ($0.170$) and OmiCLIP ($0.097$). This advantage was robust to the number of genes analyzed: as shown in Fig.~\ref{fig:fig3-he2st_main}C for colorectal, \ours\ consistently maintained a distinct lead over all three baselines across the top 100 to 2,000 variable genes.
}

\noindent
A substantial fraction of the best-predicted genes by \ours encode regulators of cell morphology and cytoskeletal dynamics. Smooth muscle--associated actin components such as \textit{MYH11}, \textit{ACTA2}, \textit{TAGLN} and \textit{CNN1} control stress fibre assembly and contractility, thereby influencing cell shape and motility\cite{owens2004molecular}. Adhesion-associated molecules such as \textit{FERMT2}, \textit{FLNA}, and \textit{TNS1} link integrins to the actin cytoskeleton, thereby regulating the turnover of focal adhesions and consequently influencing cancer cell migration\cite{parsons2010cell,hamidi2018every}. Extracellular matrix proteins such as \textit{COL1A2} further contribute to tissue organization and structural cues that shape cellular morphology\cite{frantz2010extracellular}. Several genes also converge on colorectal cancer biology: \textit{RNF43} is associated with Wnt/$\beta$-catenin signalling\cite{giannakis2014rnf43}, while altered \textit{EPCAM, CLDN3, CDH1} are linked to barrier breakdown and invasive growth\cite{tutlewska2013germline,christou2017cadherin}. Protease--inhibitor imbalance (e.g., exemplified by \textit{ST14} and SPINT2) and dysregulation of \textit{RAB25} or \textit{ELF3} link to epithelial homeostasis and metastasis\cite{roversi2018serine, nam2010loss}. These findings indicate that histology-based gene expression predictions are grounded in both morphological correlates and oncogenic programs.

\noindent
{
Fig. \ref{fig:fig3-he2st_main}D visualizes the H\&E images, ground-truth measured ST, as well as predicted ST from H\&E using \ours and Virchow. We highlight a panel of clinically relevant biomarker genes across multiple cancer types to demonstrate predictive performance.
\textit{FGFR3}, which defines a distinct molecular subtype in bladder cancer and a therapeutic target \cite{ascione2023role}, was predicted with a Pearson correlation of \(r = 0.77\) using \ours versus \(r = 0.65\) with Virchow.
\textit{ERBB2 (HER2)}, a key driver of tumor progression and established drug target in breast cancer \cite{oh2020her2}, was predicted with \(r = 0.69\) compared to \(r = 0.49\).
\textit{BCL2L1}, an anti-apoptotic gene linked to colorectal cancer progression and therapy resistance \cite{sillars2012bcl2l1}, reached \(r = 0.74\) versus \(r = 0.07\).
\textit{CDK4}, a cell-cycle regulator and target of CDK4/6 inhibitors in glioblastoma and other malignancies \cite{piezzo2020targeting}, was predicted with \(r = 0.73\) compared to \(r = 0.58\).
\textit{MET}, a receptor tyrosine kinase and actionable driver in renal cell carcinoma with approved MET inhibitors \cite{rivas2022met}, showed a marked improvement (\(r = 0.84\) vs. \(r = 0.45\)).
\textit{COL1A1}, a matrix component associated with fibrosis, tumor progression, and poor prognosis in liver cancer \cite{ma2019collagen}, achieved \(r = 0.70\) versus \(r = 0.39\).
\textit{EGFR}, a major oncogenic driver and therapeutic target in lung cancer \cite{passaro2021overcoming}, reached \(r = 0.70\) compared to \(r = 0.61\).
Finally, \textit{KLK2}, a prostate-specific protease and biomarker in prostate cancer \cite{shang2014human}, was predicted with a correlation of \(r = 0.89\) using \ours versus \(r = 0.10\) with Virchow.
}

\begin{figure*}[!h]
\centering
\includegraphics[width=0.97\linewidth]{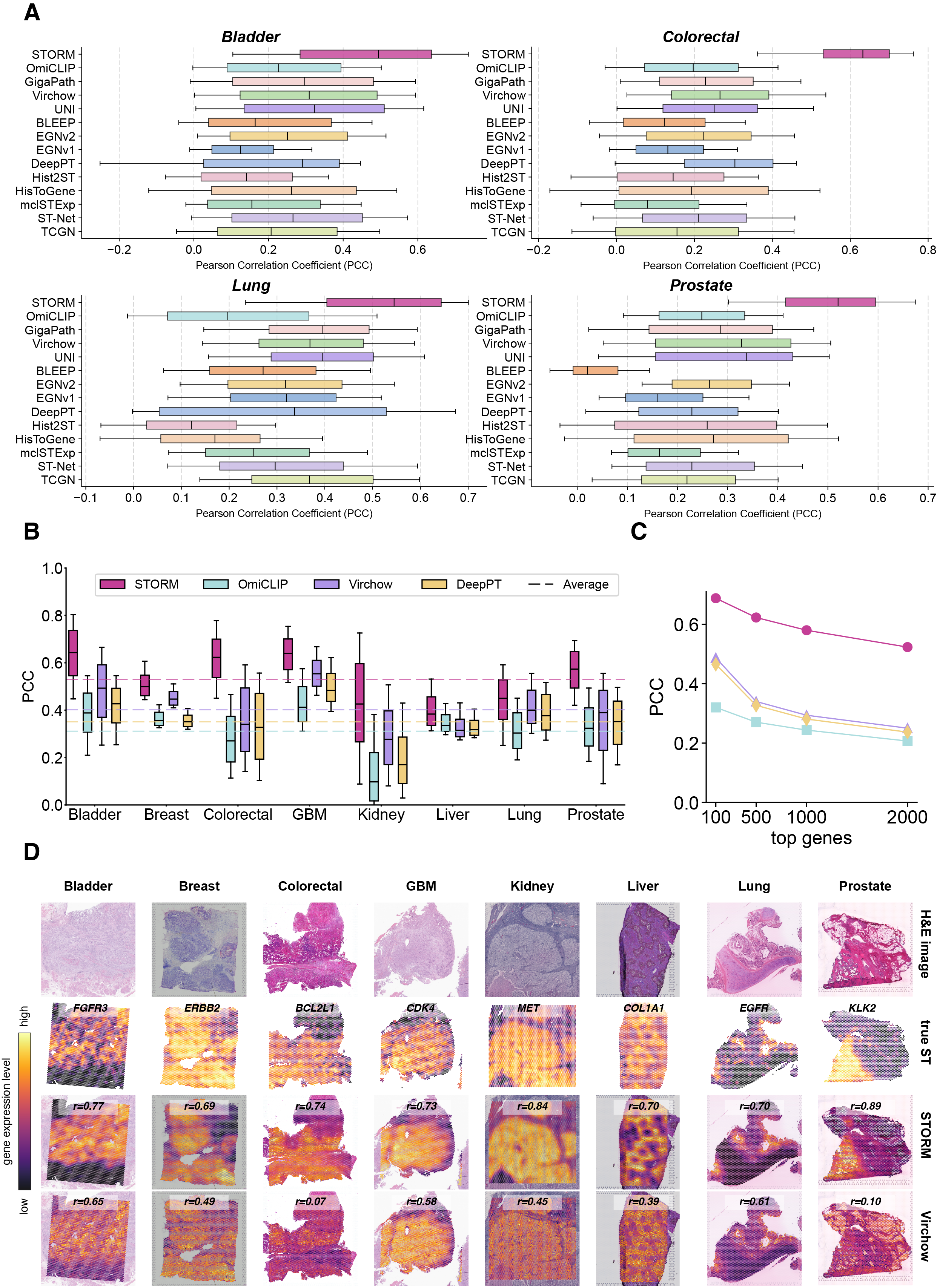}
\end{figure*}
\clearpage
\begin{figure*}[!h]
\caption{
{
\textbf{Predicting spatial transcriptomics from H\&E images.}
\textbf{(A)} Benchmark performance of \ours against competing methods for bladder, colorectal, lung, and prostate showed that \ours consistently achieved the highest Pearson correlation coefficient (PCC) in all datasets.  Boxplots show gene-wise PCC between predicted and observed gene expression for 500 HVGs. The central line represents the median; boxes span the interquartile range (IQR, 25th--75th percentile). Whiskers extend to the 10th and 90th percentiles.
\textbf{(B)}  Models are trained to predict whole transcriptomes and then we report the PCC of the top 500 genes across eight tissue types. Boxplots display the distribution of PCC values, illustrating that \ours achieves consistently higher accuracy than OmiCLIP across all tissues.
\textbf{(C)}  PCCs as a function of top gene panel size (100--2,000 genes) on the colorectal dataset. Points indicate the median PCC at each panel size.
\textbf{(D)} Visualization of spatial gene expression predicted from H\&E for clinically relevant biomarkers. Rows show, from top to bottom: H\&E image, measured gene expression, \ours predictions, and Virchow predictions. Compared with Virchow, \ours produced substantially higher correlations with ground-truth gene expression.
}
}
\label{fig:fig3-he2st_main}
\end{figure*}

\heading{{Generalization across histologies and ST platforms}}

{
\noindent
\textbf{Generalization to under-represented and unseen histologies.}
A critical challenge is the long-tailed distribution of pre-training data, where rare diseases or specific tissue subtypes are often scarce. To evaluate whether \ours\ generalizes beyond the dominant cancer types, we obtained an external validation dataset from the MOSAIC Initiative\footnote{\url{https://ega-archive.org/dacs/EGAC50000000398}} , which includes three tumor types that are under-represented or absent in the training data: Diffuse Large B-Cell Lymphoma (DLBCL, $n=10$ slides), Mesothelioma ($n=10$ slides), and Ovarian Cancer ($n=15$ slides). These datasets represent distinct histological challenges, from the dense, amorphous cellularity of lymphomas to the complex stromal interfaces of mesothelioma.

\noindent
When evaluated on these expanded datasets, \ours\ maintained high predictive fidelity, consistently outperforming both task-specific baselines and foundation models.
In Ovarian cancer, \ours\ achieved a median PCC of $0.586$, surpassing the top-performing pathology foundation model Virchow ($0.447$) by a margin of $31.0\%$. \ours\ also demonstrated a $54.6\%$ relative improvement over the best task-specific model DeepPT ($0.379$). Similarly, in Mesothelioma, \ours\ ($0.447$) demonstrated a $24.9\%$ relative improvement over Virchow ($0.358$) and effectively doubled the performance of task-specific methods like EGNv2 and DeepPT ($0.149$). The performance gap was most pronounced in DLBCL, a tissue type characterized by high cellular density and subtle morphological variations that challenge standard histological encoders. \ours\ achieved a median PCC of $0.557$ with a relative performance gain of $56.0\%$ over the second-best method OmiCLIP of $0.357$. Virchow achieved a PCC of $0.345$, while UNI dropped to a PCC of $0.151$, performing worse than several task-specific baselines (e.g., ST-Net, $0.301$). DeepPT failed to generalize in this context, achieving a PCC of only $0.072$. This dissociation suggests that while large-scale morphological pretraining (as used in UNI, Virchow, and GigaPath) confers general robustness, it is insufficient for resolving the molecular heterogeneity of lymphoid tissues. Collectively, these results confirm that \ours\ learns transferable representations that generalize to data-scarce and histologically distinct environments.

\noindent\textbf{Generalization to high-resolution ST platforms.} Although \ours\ was pretrained primarily on  10x Visium data, its architecture enables seamless adaptation to diverse spatial resolutions and capture technologies. To validate this cross-platform generalization, we evaluated the model on three distinct high-resolution technologies from the SPATCH dataset~\cite{ren2025systematic}: 10x Xenium (Ovarian cancer, 5k gene panel), 10x Visium HD (Colorectal cancer, $8~\mu\text{m}$ binning), and NanoString CosMx (Ovarian and Colorectal cancer, 1k gene panel). For each measurement location, defined as either an $8~\mu$m Visium HD bin or an individual segmented cell for Xenium and CosMx, we extracted an image patch centered at the corresponding coordinates from the H\&E image. In each case, we standardized the evaluation by predicting a panel of 500 HVGs. We use the same baselines in the Visium experiment for comparison, except for graph-based methods (Hist2ST, EGNv2, and TCGN), which are computationally intractable in high-resolution slides.

\noindent
\ours\ demonstrated robust adaptability and consistently outperformed state-of-the-art pathology foundation models and specialized ST prediction methods across all modalities. For the 10x Xenium platform, \ours\ achieved a median gene-level PCC of \textbf{0.209}, more than doubling the performance of the next-best foundation model (GigaPath: 0.088) and the second-best specialized method (STNet: 0.092).  Similarly, in the Visium HD colorectal cohort, the model successfully deconvolved complex tumor microenvironments with a median PCC of \textbf{0.298}, surpassing the runner-up Virchow (0.271). This performance extended to the NanoString CosMx platform in both ovarian (PCC = \textbf{0.295}) and colorectal (PCC = \textbf{0.280}) tissues, significantly exceeding Virchow PCC = 0.220 and PCC = 0.166, respectively. These results confirm that \ours\ learns fundamental, morphology-driven representations without overfitting to platform-specific technical artifacts.

}

\heading{Multimodal integration of H\&E and virtual ST enhances prognosis prediction}

\noindent
Having demonstrated the effectiveness of \ours for predicting spatial transcriptomics from H\&E, we next evaluated its potential for improving the prediction of clinical outcomes in large patient cohorts. Traditional deep learning models based solely on H\&E are subject to a significant risk of overfitting that results in limited generalizability. Due to their black-box nature, these models also lack biological interpretability which is crucial for high-stakes applications such as outcome prediction. Incorporating molecular information such as spatial transcriptomics may improve the model generalizability while enhancing interpretability. However, owing to its high cost and complexity, true ST is only used in research settings with limited availability and cannot be deployed to clinical practice.
{We use two multimodal baselines for comparisons: \textbf{DeepPT}, the strongest method in the task-specific category; \textbf{OmiCLIP}, the existing H\&E--ST  foundation model.}

\noindent
Here, we present a multimodal approach for prognosis prediction by integrating morphology from routine H\&E and molecular information from virtual ST. Briefly, we first applied \ours to generate virtual ST directly from H\&E slides, producing tissue spots (55~$\mu$m diameter) with histological and matched transcriptomic features. These paired representations are encoded, integrated via a cross-attention fusion module, and aggregated through an attention-based multiple instance learning (AbMIL) framework to predict patient-level prognosis. (Fig.~\ref{fig:fig4-prognosis_main}A). We trained prognosis prediction models for colorectal and non-small cell lung cancers in the PLCO dataset\cite{zhu2013prostate} (with a 4:1 split for training and cross-validation), and evaluated their generalizability on two independent test cohorts: SURGEN\cite{myles2025surgen} (colorectal) and NLST\cite{nci_cdas_nlst} (lung). Prognostic performance was assessed by C-index, using disease-specific survival (DSS) as the endpoint.

\noindent
For prognosis prediction, \ours significantly outperformed baseline models using single modality (H\&E-only and virtual ST-only), { the task-specific baseline DeepPT}, and OmiCLIP. Across the validation and test cohorts in both cancers,\ours consistently achieved the highest C-index (Table~\ref{tab:prognosis-95ci}), ranging from \textbf{0.701 to 0.741}{, while DeepPT showed lower performance across all cohorts}. The improvements over alternative methods were highly robust, with a clear separation in C-index distributions across 1,000 bootstraps (Fig.~\ref{fig:fig4-prognosis_main}B; $p<0.0001$). This is true even in settings where unimodal data were moderately informative, such as the Lung (NLST) cohort (H\&E-based model: $0.630$), multimodal integration with \ours delivered substantial further gains ($0.710$, 13\% relative improvement). Interestingly, virtual ST alone showed a very similar performance to H\&E-based models. We directly compared the prognostic performance of \ours against unimodel approaches, which achieved a positive net reclassification index of 11.1\% (colorectal) and 13.5\% (lung) over H\&E-only models for predicting 5-year DSS. Finally, \ours achieved superior performance over an existing multimodal foundation model, OmiCLIP{, and also consistently exceeded the performance of DeepPT across all cohorts}. For instance, in the CRC (SURGEN) external test cohort, \ours demonstrated a 49\% relative improvement in C-index (0.701 vs 0.472) over OmiCLIP, suggesting the strong feature representation and integration capability of \ours.

\noindent
We next evaluated whether the risk scores derived from \ours could stratify patients into clinically meaningful subgroups (Fig.~\ref{fig:fig4-prognosis_main}C). Applying a fixed threshold based on the median risk score in the validation cohort, Kaplan--Meier analyses revealed clear separation between high- and low-risk groups across independent test cohorts, with hazard ratios of 3.19 CRC (PLCO), 3.55 CRC (SURGEN), 2.89 Lung (PLCO), and 2.52 Lung (NLST), respectively. Log-rank tests showed significant differences in survival, confirming the robustness and generalizability of \ours across patient populations and cancer types.

\noindent
We performed multivariate Cox regression analysis by adjusting for established clinical risk factors such as stage, grade, age, and sex. \ours remained an independent prognostic factor in both colorectal cancer (HR~=~3.32, 95\% CI: 1.98--5.57, $p<0.001$) and lung cancer (HR~=~2.38, 95\% CI: 1.67--3.38, $p<10^{-6}$), as shown in Fig.~\ref{fig:fig4-prognosis_main} (D).
Importantly, \ours risk prediction consistently outperformed pathologic grade and stage, and integration of \ours with clinical variables further improved prognosis prediction performance (Fig.~\ref{fig:fig4-prognosis_main} (E)).
These results confirm that \ours captures distinct biological information not explained by stage alone and provides complementary prognostic insight, with the potential for refining risk stratification.

\noindent
To further explore this, we conducted subgroup analyses stratified by cancer stage, which is the dominant factor for adjuvant treatment decisions, but significant uncertainty remains. We found that \ours could further stratify patients within stage II and III colorectal cancer (HR = 2.40 and 4.85, respectively, $p < 0.05$). Notably, there were significant differences in 5-year DSS rates, 58\% vs 89\%, between the high vs low-risk groups in stage III colorectal cancer. Similar results were obtained for stage I and II non-small cell lung cancer. These results highlight the potential utility of \ours for improving decision-making on adjuvant therapy.

\begin{figure*}[!h]
\centering
\includegraphics[width=0.99\linewidth]{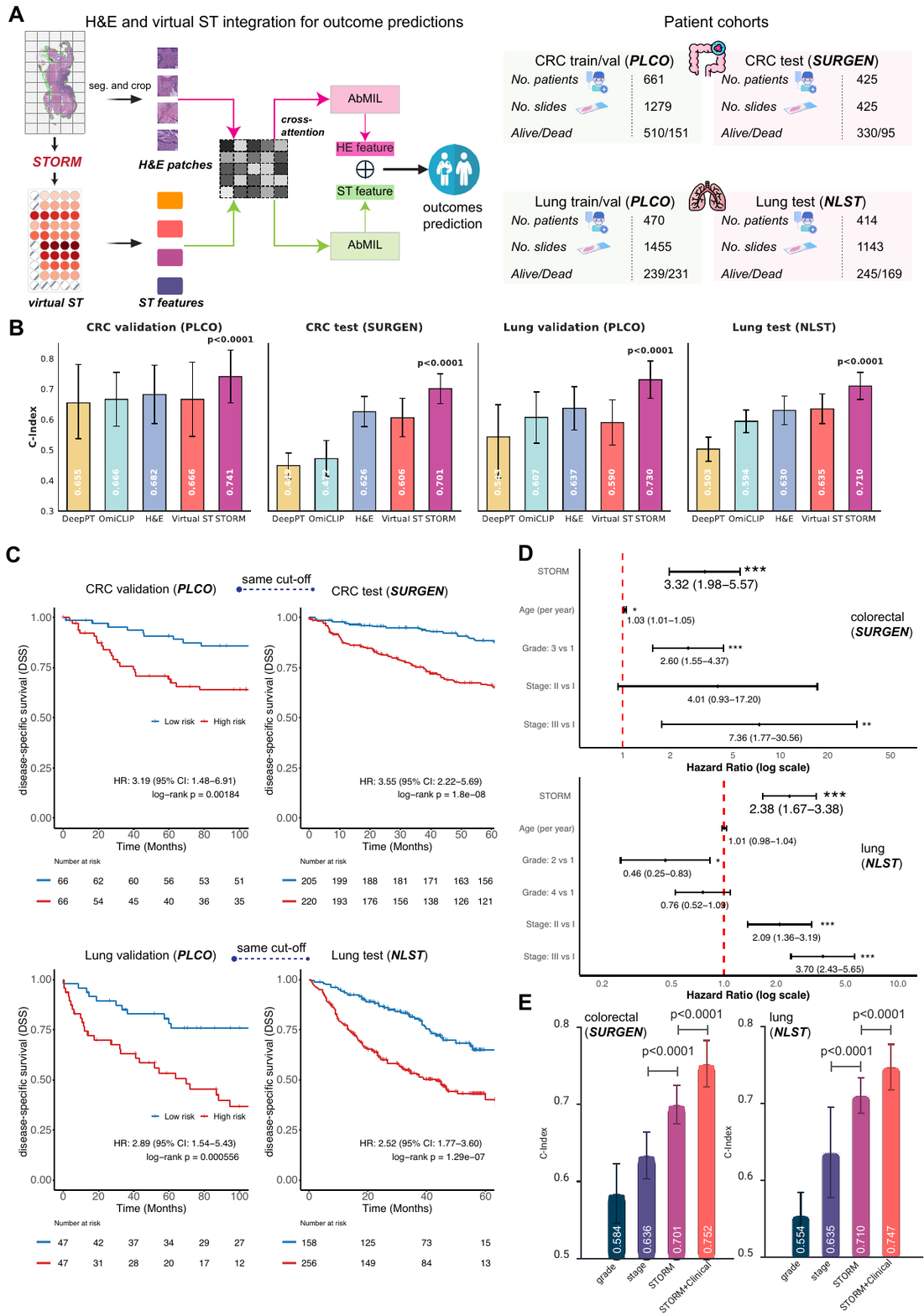}
\end{figure*}
\clearpage
\begin{figure*}[!h]
\caption{\textbf{Prognosis prediction with multimodal integration by \ours.}
\textbf{(A)} Workflow for applying \ours to prognosis prediction based on routine H\&E slides. First, \ours was used to generate virtual ST from routine H\&E slides, yielding paired H\&E images and virtual ST. Features from both modalities were encoded by \ours, integrated through a cross-attention module, and aggregated using an attention-based multiple instance learning (AbMIL) model. Final predictions were generated from the multimodal embedding via an MLP layer. We evaluated \ours using real-world large clinical cohorts of lung and colorectal cancers, with disease-specific survival (DSS) as the endpoint across training, validation, and independent test sets.
\textbf{(B)} \ours significantly outperformed unimodal baseline models (H\&E-only, virtual ST-only) and the H\&E--ST model OmiCLIP for prognosis prediction. The improvements were highly significant ($p<0.0001$). Column plots show 1,000 bootstrap samples, with bars indicating mean values and error bars representing standard deviations.
\textbf{(C)} Kaplan--Meier curves demonstrate significant patient stratification into high- and low-risk groups. A fixed cut-off value was applied to test cohorts.
\textbf{(D)} Multivariate analysis adjusting for clinical risk factors, including age, grade, and stage. In both colorectal and lung cancer, \ours remained an independent prognostic factor ($p<0.001$). In the figure, $*$ denotes $p<0.05$; $**$ denotes $p<0.01$; and $***$ denotes $p<0.001$.
\textbf{(E)} Comparison of \ours risk prediction with grade and stage. Integration of \ours with clinical variables further improved performance. Column plots show 1,000 bootstrap samples,
with bars indicating mean values and error bars representing standard deviations.
}
\label{fig:fig4-prognosis_main}
\end{figure*}

\heading{Biological interpretation of \ours for prognosis prediction}

\noindent
To aid in the interpretation of \ours model's prediction from both morphologic and molecular perspectives, we conducted a comprehensive multi-level analysis at the \textit{tissue}, \textit{cell}, and \textit{gene} levels.
To explore the biological relevance of the model predictions, we used Integrated Gradients (IG)\cite{sundararajan2017axiomatic} attributions (hereafter referred to as risk scores) to quantify feature contributions to model predictions, with positive and negative attributions indicating increased and decreased risk, respectively.

\noindent
First, we provide \textbf{tissue}-level interpretations with pathologist annotations of the model-predicted high- and low-risk image patches.
For colorectal cancer, the high-risk regions reveal: ROI 1, a micropapillary subtype (defined by small clusters of tumor cells within stromal spaces mimicking vascular channels, comprising $\geq$5\% of the tumor), which is associated with lymph node metastasis and poor prognosis; ROI 2, high-grade tumor budding, poorly differentiated clusters, and infiltrative growth, all adverse prognostic features. In addition, deep invasion into the bowel wall and beyond indicates advanced stage and poor prognosis.
In contrast, the low-risk regions show: ROI 1, a prominent peritumoral lymphocytic infiltrate (Crohn's-like reaction) associated with improved prognosis; ROI 2, a combination of peritumoral lymphocytic infiltration and a pushing rather than infiltrative growth pattern, both linked to favorable outcomes.
For lung cancer, the high-risk regions reveal: ROI 1, spread through air spaces (STAS), a marker of poor prognosis; ROI 2, large areas of coagulative tumor necrosis, also associated with adverse outcomes.
The low-risk regions show: ROI 1, a well-differentiated tumor with a papillary growth pattern and prominent tumor-infiltrating lymphocytes; ROI 2, brisk peritumoral lymphocytic infiltration, another feature predictive of favorable prognosis.

\noindent
We quantified \textbf{cell}-type composition from H\&E image patches with the top and bottom 5\% of risk scores. Using the HistoPLUS\cite{adjadj2025towards} model, we segmented and classified individual cells in image patches. Here, we focused on six major cell types in the tumor microenvironment, including cancer cells, lymphocytes, macrophages, fibroblasts, plasma cells, and endothelial cells.
Significant differences in cellular composition were observed between prognostic groups. As expected, cancer cells were highly enriched in high-risk patients ($0.50$ vs.\ $0.21$, $p = 1.1 \times 10^{-97}$). Stromal and vascular components further distinguished risk groups: fibroblasts were significantly elevated in high-risk regions ($0.36$ vs.\ $0.26$, $p = 4.3 \times 10^{-25}$), and endothelial cells were modestly higher ($0.15$ vs.\ $0.12$, $p = 0.022$).  In contrast, there was abundant immune infiltration in low-risk patients including lymphocytes ($0.46$ vs.\ $0.26$, $p = 5.2 \times 10^{-97}$), plasma cells ($0.26$ vs.\ $0.18$, $p = 1.3 \times 10^{-8}$), and macrophages ($0.13$ vs.\ $0.10$, $p = 9.5 \times 10^{-5}$). Taken together, these findings suggest that the model identified tumor-rich, fibroblast- and endothelium-enriched regions in high-risk groups, while regions with immune cell infiltration with adaptive and innate immunity contributed to low-risk predictions.

\noindent
We further explored \textbf{gene} level interpretations using the lung cancer cohort. Similarly to above, we computed risk scores using the IG attributions method for each gene across all slides, and identified a coherent set of outcome-associated genes. Many of these genes including \textit{CD74}, \textit{HLA-A/C}, \textit{B2M}, \textit{HSP90AA1/AB1}, \textit{EPAS1}, \textit{EZR}, and \textit{XPO1} are well-established regulators of tumor progression with prognostic significance. We next conducted gene set enrichment analysis of the top 50 genes, and identified several enriched molecular pathways (FDR $<0.05$ ). These include Interferon-Gamma Response, Unfolded Protein Response, G2--M Checkpoint, PI3K/AKT/mTOR signalling, TGF-$\beta$ signalling, Mitotic Spindle, TNF-$\alpha$ signalling via NF-$\kappa$B, Hypoxia, Myc Targets V1 and IL-6/JAK/STAT3 signalling. These results demonstrate that \ours revealed a prognostic gene set at the intersection of proliferative, stress-response and immunomodulatory pathways, highlighting its potential for identifying candidates for combination therapeutic targeting.

\begin{figure*}[!h]
\centering
\includegraphics[width=1.0\linewidth]{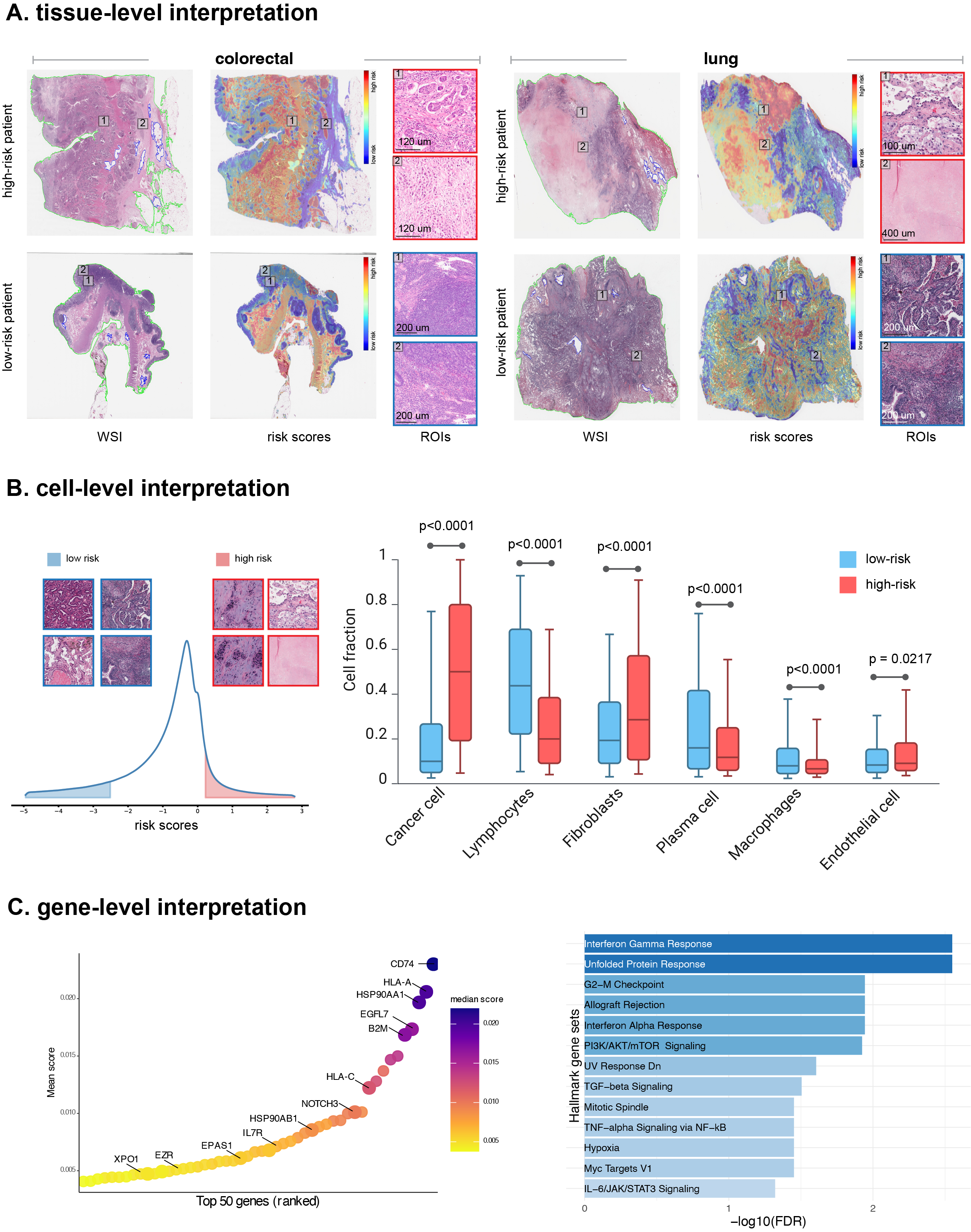}
\end{figure*}
\clearpage
\begin{figure*}[!h]
\caption{\textbf{Biological interpretation of prognosis prediction at the tissue, cell, and gene levels.}
\textbf{(A)} Tissue-level interpretation. Representative high- and low-risk colorectal and lung cancer ROIs from the validation set. High-risk colorectal cancer ROIs show classic adenocarcinoma features, while high-risk lung cancer ROIs are enriched with extracellular mucin characteristic of mucinous adenocarcinoma. Low-risk colorectal cancer ROIs display a strong TIL response, and low-risk lung cancer ROIs exhibit chronic inflammation, hemorrhage, and fibrotic stroma, all markers of a favorable prognosis.
\textbf{(B)} Cell-level interpretation. Boxplots show the distribution of cell-type fractions across patients for cancer cells, lymphocytes, fibroblasts, plasma cells, macrophages, and endothelial cells. High-risk predictions were associated with enrichment of cancer cells, fibroblasts, and endothelial cells, whereas low-risk predictions were associated with increased lymphocytes and plasma cells. Macrophages were modestly enriched in low-risk patients. $p$-values from Welch's t-test are indicated.
\textbf{(C)} Gene-level interpretation. Risk scores from \ours identified prognostic genes (\textit{CD74, HLA-A/C, B2M, HSP90AA1/AB1, EPAS1, EZR, XPO1}) and enriched pathways (e.g., Interferon-Gamma Response, Unfolded Protein Response), highlighting proliferative, stress-response, and immune programs with therapeutic relevance.
}
\label{fig:fig5-prognosis_interpret}
\end{figure*}

\Heading{Discussion}

\noindent
In this study, we present a multimodal foundation model \ours that integrates histopathology and spatial transcriptomics to generate spatially resolved molecular insights at scale. Trained on 1.2 million spots from 632 paired H\&E--ST sections across 18 organs, \ours consistently and significantly outperformed state-of-the-art methods in spatial domain discovery, spatial gene expression prediction, as well as patient prognosis and immunotherapy response prediction across 11 independent cohorts with $>2500$ patients. Beyond predictive accuracy, \ours also revealed spatial gene expression patterns associated with prognosis and immunotherapy response, offering valuable insights into disease mechanisms and potential therapeutic targets. These findings highlight the potential of \ours to dissect the spatial organization of complex tissue architectures directly from routine histology, thereby expanding the reach of spatial biology into clinical contexts.

\noindent
By integrating ST and H\&E, \ours revealed biologically coherent spatial domains, including metabolically rewired and immune-enriched tumor niches linked to poor prognosis, and resolved immune substructures such as tertiary lymphoid regions and infiltrate subtypes with distinct functional states. These findings underscore that multimodal embeddings not only improve clustering accuracy but also uncover hidden heterogeneity within histologically uniform regions. In particular, the delineation of tumor subdomains with distinct metabolic, inflammatory, and hypoxic signatures highlights how subtle spatial variations can translate into prognostic differences, offering new avenues for therapeutic targeting. Likewise, the separation of plasma cell--dominated and Tfh/DC/NK-enriched immune niches demonstrates the model's ability to resolve fine-grained immunological circuits with implications for immunotherapy design. More broadly, the ability of \ours to integrate histological context with molecular states establishes a framework for precision oncology, where spatially grounded signatures may guide patient stratification and inform interventions tailored to microenvironmental architecture. Together, these results position \ours as a powerful tool to bridge morphology and molecular context, enabling domain-level discoveries that inform both mechanistic understanding and clinical translation.

\noindent
Accurate prediction of treatment response and prognosis has significant implications for precision medicine. In stage~II colorectal cancer, adjuvant chemotherapy is controversial and usually reserved for patients with poor prognostic features, whereas in stage~III colorectal cancer, adjuvant chemotherapy is a standard of care, but questions remain on how to best tailor therapy duration to balance efficacy with toxicity\cite{grothey2018duration}. Similarly, for early-stage non-small cell lung cancer, there is significant uncertainty around the optimal use of adjuvant therapy. Our \ours model better predicts prognosis compared with staging and can further stratify patients within each disease stage. Importantly, this is achieved with H\&E slides that are part of routine clinical practice. If validated, it can be used to refine risk stratification and guide adjuvant therapy.

\noindent
In conclusion, we present a H\&E-ST foundation model \ours that integrates histology and spatial transcriptomics. By uniting the accessibility of histology with the molecular depth of ST, \ours establishes a scalable framework for spatially informed biological discovery and accelerates translation into clinical practice.

\Heading{Online Methods}





\heading{\ours model design and pretraining strategy}

\noindent The architecture of \ours follows a two-level hierarchical design. \textit{At the spot level}, pretrained encoders independently extract modality-specific representations from H\&E patches and ST spots. \textit{At the spatial level}, features from a $5\times5$ neighborhood of spots are jointly processed, enabling the model to capture local tissue organization and broader spatial context.

\noindent
\textbf{Spot-level encoder.}
For a spot with gene expression $\boldsymbol{X} \in \mathbb{R}^{1 \times M}$, a pretrained ST foundation model (NOVAE\cite{blampey2024novae}) produces embedding $\boldsymbol{H}_G = f_{\mathrm{ST}}(\boldsymbol{X}) \in \mathbb{R}^{1 \times h}$ ($h=64$). For the corresponding H\&E patch $\boldsymbol{V} \in \mathbb{R}^{1 \times C}$, a pretrained pathology foundation model (H0-mini\cite{filiot2025distilling}) produces embedding $\boldsymbol{H}_P = f_{\mathrm{HE}}(\boldsymbol{V}) \in \mathbb{R}^{1 \times d}$ ($d=768$).

\noindent
\textbf{Spatial encoder.} The spatial encoder integrates multimodal information across the spot neighborhood. Each block contains a shared multi-head self-attention (MSA) module and two modality-specific feed-forward networks (modality experts). Tokens are routed to the appropriate expert based on modality, while the shared MSA aligns features across modalities. This unified architecture supports both multimodal fusion (H\&E + ST) and unimodal inference (H\&E only): when ST inputs are absent, the attention operation collapses to standard H\&E self-attention.

\noindent Input embeddings combine feature tokens, a learnable modality embedding, and positional embeddings encoded via Attention with Linear Biases (ALiBi)\cite{press2021train}:
\begin{equation}
    \boldsymbol{H}_{0,i} = \boldsymbol{H}_i + \boldsymbol{M}_i + \boldsymbol{P}_i,
\end{equation}
where $\boldsymbol{H}_i \in \{\boldsymbol{H}_{P,i}, \boldsymbol{H}_{G,i}\}$ depending on modality, and $\boldsymbol{M}_i \in \{\boldsymbol{M}_H, \boldsymbol{M}_G\}$ is the modality embedding. Each spatial block updates representations as:
\begin{equation}
    \boldsymbol{H}_l^{\prime} = \operatorname{MSA}(\operatorname{LN}(\boldsymbol{H}_{l-1})) + \boldsymbol{H}_{l-1}, \quad
    \boldsymbol{H}_l = \operatorname{MoME\text{-}FFN}(\operatorname{LN}(\boldsymbol{H}_l^{\prime})) + \boldsymbol{H}_l^{\prime}.
\end{equation}

\noindent
\textbf{Decoders and pretraining.}
We adopt a unified masked data modeling strategy in which H\&E--ST spots are treated as ``words'' and sequences of neighboring spots as ``sentences.'' Randomly masked spots are reconstructed using lightweight modality-specific decoders (an MLP followed by two Transformer blocks), discarded after pretraining. We fix the number of visible spots to 5 per modality (one-fifth of the $5\times5$ window). The pretraining loss minimizes negative log-likelihood over masked positions:
\begin{equation}
\mathcal{L}_{\text{pretrain}} =
-\frac{1}{|\mathcal{M}_{\mathrm{HE}}|}\sum_{i \in \mathcal{M}_{\mathrm{HE}}} \log p_{\mathrm{HE}}\!\left(z_{\mathrm{HE},i}\mid \mathbf{X}^{\mathcal{M}}\right)
-\frac{1}{|\mathcal{M}_{\mathrm{ST}}|}\sum_{i \in \mathcal{M}_{\mathrm{ST}}} \log p_{\mathrm{ST}}\!\left(z_{\mathrm{ST},i}\mid \mathbf{X}^{\mathcal{M}}\right).
\end{equation}

\noindent
\textbf{Pretraining settings.}
Models are pretrained for 1000 epochs on 1.2M spots using AdamW (lr $=10^{-4}$, weight decay $=0.05$), with cosine decay after a 40-epoch warmup. Batch size is 1024; training uses 8 NVIDIA H100 GPUs with automatic mixed precision. The ST encoder (NOVAE) is frozen; the H\&E encoder (H0-mini) and spatial encoder are trainable with layer-wise learning rate decay ($\lambda=0.7$). H\&E augmentations include random resized cropping and flips; no augmentations are applied to ST data. For gene panel heterogeneity (Xenium: $<$500 genes; Visium: $>$10{,}000), the decoder predicts the union of all gene panels but masks genes absent from a given input.

\heading{Ablation study of \ours design}

\noindent We conducted a systematic ablation study holding all settings fixed except the component under evaluation, across 11 tissue datasets using identical splits, preprocessing, and optimization.
\textbf{H\&E encoder}: we compared replacing the pretrained H0-mini with (i) a randomly initialized model and (ii) H0-mini used directly without \ours.
\textbf{ST encoder}: we replaced NOVAE with raw highly variable gene vectors.
\textbf{Spatial encoder}: we evaluated processing each spot independently (grid size $=1$, no spatial context).
\textbf{Pretraining}: we compared the full pretrained model against training from random initialization on the downstream task only.

\heading{\ours application: spatial domain discovery}

\noindent \ours is applied directly to multimodal embeddings without additional training. For inference, neighboring $5\times5$ spots are cropped from co-registered H\&E and ST slides and embedded by \ours. These features are clustered via K-means and evaluated against pathologist annotations using eight metrics: NMI, ARI, FMI, Homogeneity (HOM), Completeness (COM), CHAOS, PAS, and ASW. Differentially expressed genes across spatial domains were identified using the Wilcoxon rank-sum test (Scanpy\cite{wolf2018scanpy}), with Benjamini--Hochberg correction; over-representation analysis used the MSigDB hallmark gene set collection\cite{subramanian2005gene} (Fisher's exact test, FDR $<0.05$).

\heading{\ours application: predicting Visium ST from H\&E}

\noindent A lightweight MLP prediction head is attached to \ours for H\&E-to-ST inference. We evaluate on eight organ datasets: bladder ($n=15$), breast ($n=5$), colorectal ($n=37$), GBM ($n=10$), kidney ($n=24$), liver ($n=8$), lung ($n=20$), and prostate ($n=23$), using an 80/20 slide-level train/test split. We benchmarked against 13 methods: task-specific models (BLEEP\cite{xie2023spatially}, mclSTExp\cite{min2024multimodal}, DeepPT\cite{hoang2024deep}, EGNv1\cite{yang2023exemplar}, EGNv2\cite{yang2024spatial}, Hist2ST\cite{zeng2022spatial}, ST-Net\cite{he2020integrating}, HisToGene\cite{Pang2021LeveragingII}, TCGN\cite{xiao2024transformer}), pathology foundation models (GigaPath\cite{xu2024whole}, UNI\cite{chen2024towards}, Virchow\cite{vorontsov2024foundation}), and H\&E-ST foundation model OmiCLIP\cite{chen2025visual}. Performance is reported as gene-wise Pearson correlation coefficient (PCC) over 500 HVGs. During fine-tuning, only the prediction head is trained (AdamW, lr $=5\times10^{-4}$, 10 epochs, cosine decay). The fine-tuning loss is:
\begin{equation}
    \mathcal{L}(\boldsymbol{Y}, \hat{\boldsymbol{Y}}) = \frac{1}{NG} \sum_{i=1}^N \sum_{j=1}^G \left( \lambda_1 \left| y_{ij} - \hat{y}_{ij} \right| + \lambda_2 (y_{ij} - \hat{y}_{ij})^2 \right).
\end{equation}

\heading{\ours application: predicting single-cell ST from H\&E}

\noindent We use the SPATCH benchmark\cite{ren2025systematic} for single-cell ST prediction, comprising Visium HD FFPE ($8~\mu$m bins, one colorectal slide), CosMx 6K (one ovarian and one colorectal slide), and Xenium 5K (one ovarian slide), all at $40\times$ magnification. After platform-aware quality control and normalization, analyses are restricted to 500 shared highly variable genes with a spatially contiguous 80/20 x-coordinate split. We benchmarked against 10 methods using the same evaluation metrics as for Visium prediction, with the local context set to 9 neighbors per spot (vs.\ $5\times5$ for Visium).

\heading{\ours application: prognosis}

\noindent \ours is evaluated in two clinical settings: survival prognosis and immunotherapy response prediction. H\&E WSIs are preprocessed into $55\,\mu$m patches ($110\times110$ px at $20\times$; $220\times220$ px at $40\times$, resized to $224\times224$), and virtual ST is inferred using the fine-tuned \ours encoder. For outcome prediction, patch-level H\&E and virtual ST features are compressed with Perceivers\cite{jaegleperceiver}, fused via cross-attention, and aggregated with modality-specific attention-based pooling to produce a slide-level risk score.

\noindent
\textbf{Prognosis cohorts.} For colorectal cancer (CRC), we used 1,086 patients: PLCO ($n=661$, training/internal validation) and Surgen ($n=425$, independent test)\cite{myles2025surgen}. For lung cancer, we used 884 patients: PLCO ($n=470$) and NLST ($n=414$, independent test)\cite{nci_cdas_nlst}. Virtual ST was predicted for 1,196 OncoKB cancer-relevant genes. For pancancer analysis, we utilized TCGA-CDR across 12 cancer types (4,710 patients) evaluated by 5-fold cross-validation. Baselines included DeepPT and OmiCLIP, all using the same H\&E and virtual ST multimodal inputs.

\noindent

\noindent
\textbf{Biological interpretation.} Patch-level importance was computed using Integrated Gradients (IG)\cite{sundararajan2017axiomatic} with a zero baseline, normalized by min--max scaling across the WSI. For immunotherapy, a GMM classified regions into high- and low-risk zones; gene perturbation scores (fold changes) were aggregated across slides via Robust Rank Aggregation (RRA), and top-ranked genes were analyzed with GO and Reactome pathway enrichment\cite{milacic2024reactome}.

\heading{Statistical analysis}
For tasks with independent test sets, performance variation was assessed by non-parametric bootstrapping (1,000 samples). For cross-validation tasks, mean and standard deviation were estimated across folds. Statistical significance was assessed using the two-sided Mann-Whitney U test, Welch's t-test, or two-sided Wilcoxon signed-rank test as indicated in figure captions. Survival analysis used the concordance index (c-index); Kaplan--Meier curves were generated using median predicted risk as the cutoff, with log-rank testing for group comparison.

\section*{Data Availability}

\noindent
For model pretraining, we used publicly available spatial transcriptomics cohorts derived from \textit{Homo sapiens} tissues, spanning a wide range of organs and disease states. Representative examples include FFPE human skin primary dermal melanoma profiled with Xenium (\url{https://www.10xgenomics.com/datasets/xenium-primary-dermal-melanoma}); FFPE human prostate adenocarcinoma with Xenium (\url{https://www.10xgenomics.com/datasets/xenium-prostate-adenocarcinoma}); multiple colorectal cancer and healthy bowel samples profiled with Visium HD and Visium, such as the COLON MAP cohort (\url{https://humantumoratlas.org/explore?selectedFilter=hta11}); human cardiac tissue profiled with Visium as part of the Heart Cell Atlas (\url{https://www.heartcellatlas.org/}); image-based spatial transcriptomics datasets of healthy and diseased lung available via GEO (\url{https://www.ncbi.nlm.nih.gov/geo/query/acc.cgi?acc=GSE}); FFPE human ovarian, pancreatic, lung, and brain cancers with immune-oncology panels profiled on Xenium (\url{https://www.10xgenomics.com/datasets}); and FFPE human lymph node, lymphoid, bone, and eye tissues with corresponding datasets on the 10x Genomics portal.

\noindent
Lung and kidney samples with H\&E images, 10x Visium ST data, and pathologist annotations for domain discovery are available at \url{https://zenodo.org/records/14620362}, while human HER2-positive breast tumor ST data can be accessed at \url{https://github.com/almaan/her2st}. For H\&E--ST prediction, we used the following datasets: 5 slides of breast cancer from 10x Genomics (\url{https://www.10xgenomics.com/datasets/human-breast-cancer-block-a-section-2-1-standard-1-1-0}); 37 slides of colorectal cancer from the Human Tumor Atlas Network (\url{https://data.humantumoratlas.org/center/hta11});  24 slides of kidney cancer from the Gene Expression Omnibus (GEO; \url{https://www.ncbi.nlm.nih.gov/geo/query/acc.cgi?acc=GSE175540}); 8 slides of liver cancer from GEO (\url{https://www.ncbi.nlm.nih.gov/geo/query/acc.cgi?acc=GSE223561}); 20 slides of lung cancer from the European Nucleotide Archive (ENA; \url{https://www.ebi.ac.uk/ena/browser/view/PRJEB52292}); and 23 slides of prostate cancer from EGA (\url{https://ega-archive.org/studies/EGAS00001006124}).
{
We also used Visium H\&E-ST datasets hosted under the MOSAIC Window Data Access Committee (DAC) with the ID EGAC50000000398: \url{https://ega-archive.org/dacs/EGAC50000000398}. Specifically, this includes datasets for 15 slides of bladder cancer (EGAD50000001251), 10 slides of glioblastoma (EGAD50000001352), 10 slides of diffuse large B-cell lymphoma (EGAD50000001702), 15 slides of ovarian cancer (EGAD50000001704), and 10 slides of mesothelioma (EGAD50000001706). }

{
\noindent
We use the SPATCH dataset for single-cell spatial transcriptomics prediction, which can be accessed at \url{https://spatch.pku-genomics.org/#/homepage}. SPATCH is a comprehensive benchmark comprising paired H\&E and ST data across multiple high-throughput platforms with subcellular or cellular resolution, including Visium HD FFPE (one colorectal cancer slide), CosMx 6K (one ovarian cancer slide and one colorectal cancer slide), and Xenium 5K (one ovarian cancer slide). }

\noindent
For colorectal cancer prognosis, we used H\&E images with follow-up labels and clinical information from two datasets: PLCO (\url{https://cdas.cancer.gov/datasets/plco/22/}) and SURGEN (\url{https://github.com/CraigMyles/SurGen-Dataset}). For lung cancer prognosis, we used H\&E images with follow-up labels and clinical information from PLCO (\url{https://cdas.cancer.gov/datasets/plco/21/}) and NLST (\url{https://www.cancerimagingarchive.net/collection/nlst/}).

\noindent
Data for all immunotherapy cohorts were collected from Stanford Hospital and are not publicly available due to patient privacy concerns. These data may be shared upon reasonable request to the corresponding author, subject to review of a brief research proposal and execution of a data use agreement.

\Heading{Competing interests}

\noindent
The authors declare no competing interests.

\Heading{Code Availability}

\noindent \textbf{Code}: The code and model are provided as {supplementary software} for peer review purposes. The code will be made publicly available for non-commercial academic research upon publication.

\end{spacing}

\clearpage

\begin{nolinenumbers}
\Heading{References}

\vspace{2mm}

\begin{spacing}{0.9}
\bibliographystyle{naturemag}
\bibliography{main}
\end{spacing}
\end{nolinenumbers}
\clearpage

\clearpage

\end{document}